\pdfoutput=1							
\documentclass[10pt, a4paper]{article}		
\usepackage{lrec2022} 
\usepackage{multibib}
\newcites{languageresource}{Language Resources}
\usepackage{graphicx}
\usepackage{tabularx}
\usepackage{makecell}
\usepackage{multirow}
\usepackage{xspace}
\usepackage{scrextend}

\usepackage{soul}
\usepackage{titlesec}
\titleformat{\section}{\normalfont\large\bfseries\center}{\thesection.}{1em}{}
\titleformat{\subsection}{\normalfont\SmallTitleFont\bfseries\raggedright}{\thesubsection.}{1em}{}
\titleformat{\subsubsection}{\normalfont\normalsize\bfseries\raggedright}{\thesubsubsection.}{1em}{}
\renewcommand\thesection{\arabic{section}}
\renewcommand\thesubsection{\thesection.\arabic{subsection}}
\renewcommand\thesubsubsection{\thesubsection.\arabic{subsubsection}}
\usepackage{textcomp}

\usepackage{epstopdf}
\usepackage[utf8]{inputenc}
\usepackage[T2A,T1]{fontenc}
\usepackage[russian,greek,main=english]{babel}
\usepackage{hyperref}
\usepackage{xstring}
\usepackage{color}
\newcounter{notecounter}
\newcommand{\enotesoff}{\long\gdef\enote##1##2{}}

\enotesoff

\usepackage{microtype}
\usepackage{todonotes}
\usepackage{array,amsmath}
\usepackage{amssymb}
\usepackage{times}
\usepackage{latexsym}
\usepackage{hyperref}
\usepackage{float}
\restylefloat{table}
\usepackage{tabularx}
\usepackage[inline]{enumitem}
\usepackage{algorithm,algpseudocode}
\newfloat{algorithm}{t}{lop}
\newcolumntype{Y}{>{\centering\arraybackslash}X}
\def\methodb{CLC-B\xspace}
\def\methodn{CLC-BN\xspace}
\newcommand{\secref}[1]{\StrSubstitute{\getrefnumber{#1}}{.}{ }}

\newcommand\blfootnote[1]{%
	\begin{NoHyper}
		\begingroup
		\renewcommand\thefootnote{}\footnote{#1}%
		\addtocounter{footnote}{-1}%
		\endgroup
	\end{NoHyper}
}

\usepackage{textcomp}

\usepackage[symbol]{footmisc}

\title{Towards a Broad Coverage Named Entity Resource:\\
A Data-Efficient Approach for Many Diverse Languages}

\name{Silvia Severini, Ayyoob Imani, Philipp Dufter$^*$, Hinrich Sch{\"u}tze} 
\address{Center for Information and Language Processing \\
        LMU Munich, Germany \\
         silvia@cis.uni-muenchen.de\\}
\abstract{
Parallel corpora are ideal for extracting a \emph{multilingual named
entity (MNE) resource}, i.e., a dataset of
names translated into multiple languages. 
Prior work on extracting MNE datasets from parallel corpora required resources
such as large monolingual corpora or word aligners that 
are unavailable or perform poorly for underresourced languages. 
We
present \methodn, a new method for creating an  MNE
resource, and apply it to the Parallel Bible Corpus,
a corpus of more than 1000 languages. \methodn
learns a neural transliteration model from
parallel-corpus statistics, without requiring any other bilingual
resources, 
word aligners, or seed data. Experimental results show that \methodn
clearly outperforms prior work.
We release an MNE resource for 1340 languages and
demonstrate its effectiveness  in two downstream tasks: 
knowledge graph augmentation and bilingual lexicon induction.    \\ \newline  }

\def\tabref#1{Table~\ref{tab:#1}}

\def\secref#1{\S\ref{sec:#1}}

\begin{document}

\maketitleabstract

\blfootnote{$^*$Now at Apple.}

\section{Introduction}
Of the thousands of languages in
the world, 
a very small portion is covered by language technologies 
\cite{joshi2020state}.
\newcite{bird2020decolonising} suggests a number of 
approaches to develop 
such
technologies for low-resource
languages.  

In this paper, our goal is to create
a \emph{multilingual named entity (MNE) resource} -- by
which we mean a dataset of names translated into multiple
languages -- for a large number of low-resource languages,
in total more than a thousand.  Named entities (NEs) are
crucial for many language technologies and NLP applications,
including text comprehension, question answering,
information retrieval and relation extraction.  In this paper, we
demonstrate the effectiveness of our MNE
resource in two downstream tasks: knowledge graph
augmentation and bilingual lexicon induction.

We extract our MNE resource from the Parallel Bible Corpus
(PBC) \citelanguageresource{mayercreating2014}, a multiparallel corpus that covers more than 1300
languages.   (Note however that we do not
use Bible-specific features; therefore, our work is in
principle applicable to any parallel corpus.)
For some languages, PBC is the only available text
\citelanguageresource{wu2018creating}.
Multiparallel corpora contain sentence-level
parallel text in more than two languages. Apart from PBC,
JW300 \citelanguageresource{agic2019jw300} and
Tatoeba\footnote{\url{https://tatoeba.org}} are two other
examples of such corpora.  While the amount of data
per language provided by highly multiparallel corpora is
usually small, this type of corpus plays an important part
in compiling resources for low-resource languages.

\begin{figure}[!t]
	\centering
	\includegraphics[width=0.6\textwidth, trim={1.5cm 4cm 2cm 2cm}, clip]{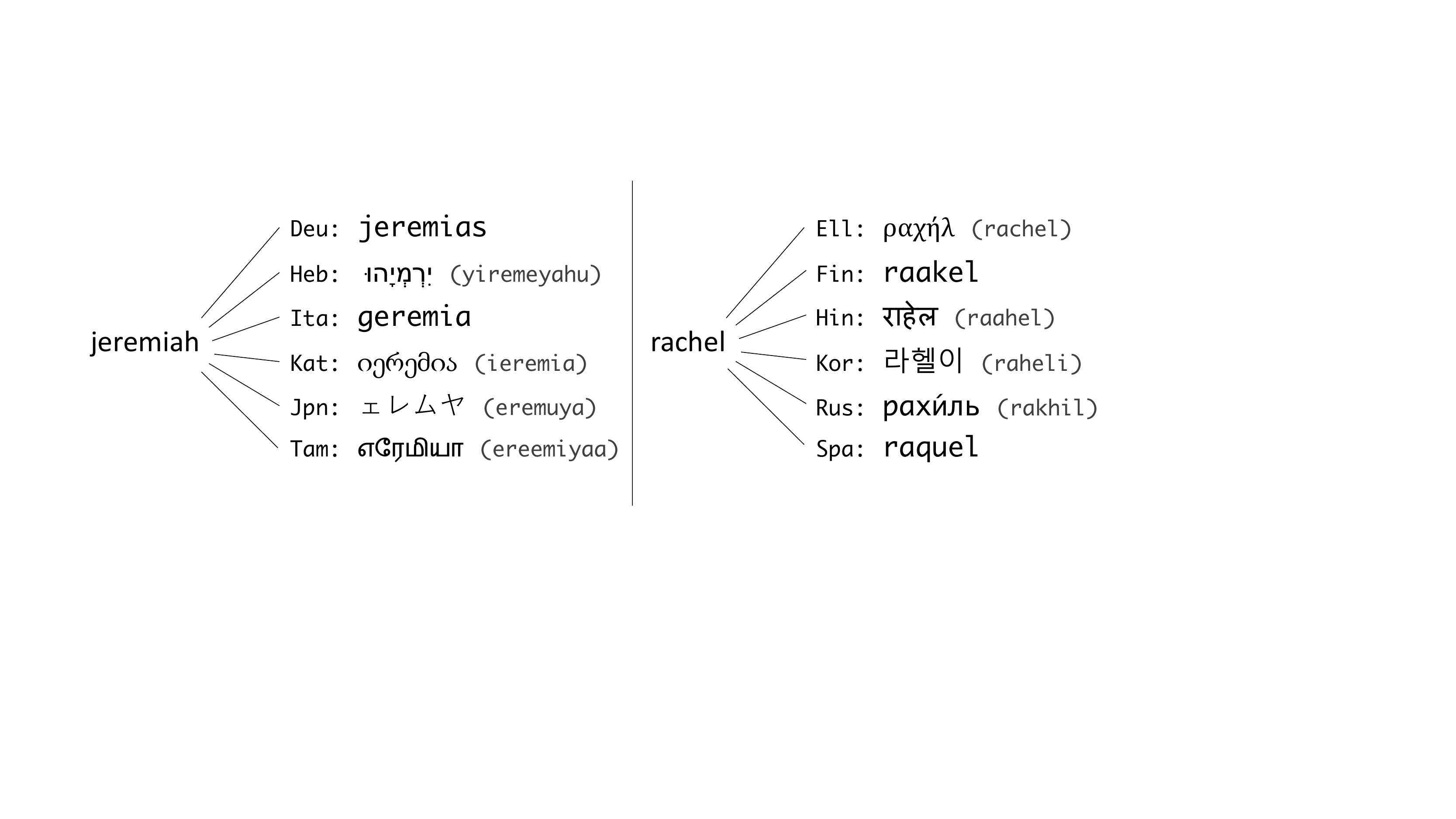}
	\vspace*{-15mm}
	\caption{Two NEs from our resource, each with a
	sample of
translations in six
different languages\label{fig:sentences}}
\end{figure}

Creating a named entity resource is comparatively easy if
sufficiently high-quality resources are available for a language. Such
resources include named entity recognizers \cite{yadav2018survey,li2020survey};
large monolingual corpora, which can be used to learn
high-quality word embeddings or high-quality contextualized
embeddings; parallel corpora that consist of large corpora (millions of
words) per language
\cite{lample2016neural,ma2016end,dasigi2011named};
or high-quality annotated data, e.g., training sets for
named entity recognition \cite{wang2014cross,wu2021unitrans,wu2020enhanced}
or implicit high-quality
annotations like hyperlinks in Wikipedia \cite{tsai2016cross}.
Recent work \cite{wu2021unitrans,li2021cross} with multilingual pretrained language models (PLMs) like BERT and 
XML-R for named entity recognition is promising, but also
relies on moderately large monolingual corpora (e.g., a
Wikipedia of decent size) to learn good quality
contextualized representations.
However, these monolingual corpora
exist only for about 100 or so languages. 
For instance, Zulu is not included but we cover it in our experiments.

In this work, our goal is to cover the large number of
languages for which these resources do not exist: no named
entity recognizers, no large monolingual (or parallel)
corpora, no annotated data (not even implicitly annotated)
and no pretrained language models (due to the lack of large
monolingual corpora).

Many low-resource languages are covered in the
PBC which gives us a chance to create resources for
languages that currently do not have any -- perhaps apart
from an entry in the World Atlas of Language Structures 
\citelanguageresource{wals}
that is too abstract for most purposes in computational linguistics.

Since PBC is a parallel corpus, the question of why we do not use word alignment naturally arises. 
However, our experiments with word
alignment on PBC were not successful for named entities. The
reason is that word alignment performance deteriorates when
parallel text is scarce \cite{och2003systematic}, especially
for named entities as most are rare words.  Our approach
therefore does not depend on a word aligner and works well
even when only a small parallel corpus is available. We
directly compare with prior work that relies on word alignment.

Based on this motivation, we introduce \methodn (\textit{Character Level Correspondence 
Bootstrapping and Neural transliteration}),
 a method for extracting
a multilingual named entity resource from a parallel corpus,
including in low-resource settings in which the available
text per language in the corpus is small.
\methodn learns a neural transliteration model from
parallel-corpus statistics, without requiring any other bilingual
resources, 
word aligners or seed data.
In the first step,
the method identifies NE correspondences in the parallel text.
It 
then learn a neural transliteration model from these
(noisy) NE correspondences.
Finally, we use the learned model to identify high-confidence NE
pairs in the parallel text.
The first step (identifying NE correspondences) works at the character-ngram level, hence 
it is applicable to languages for which a tokenizer is not available, as opposed to word alignment based approaches.
We will show that our method performs well for untokenized Japanese text.

In summary, our contributions are:
\begin{enumerate}
    \item  We present \methodn, a method that first
    identifies named entity correspondences in a parallel
    corpus and then learns a neural transliteration model
    from them.
    \item  We annotate a set of NEs to evaluate \methodn's performance on 13
			languages through crowdsourcing and show a clear
			performance increase in comparison to prior work.
			We release the gold annotated sets as a resource for future work.\footnote{\label{note1}\url{http://cistern.cis.lmu.de/ne_bible/}}
    \item Using \methodn, we create and release a named entity resource containing 674,493 names across 1340 
    languages, 503 names per language on average.\footref{note1}
\item For many languages, ours is the first published resource. 
    We believe that
it can be useful for future work in computational
linguistics on the more than 1000 languages covered.
    We show experimentally that this is the case for 
knowledge graph augmentation and bilingual lexicon induction.   
\end{enumerate}

\begin{figure*}[!t]
	\centering \includegraphics[width=.98\textwidth,
	trim={1cm 1.2cm 6cm 3.9cm},
	clip]{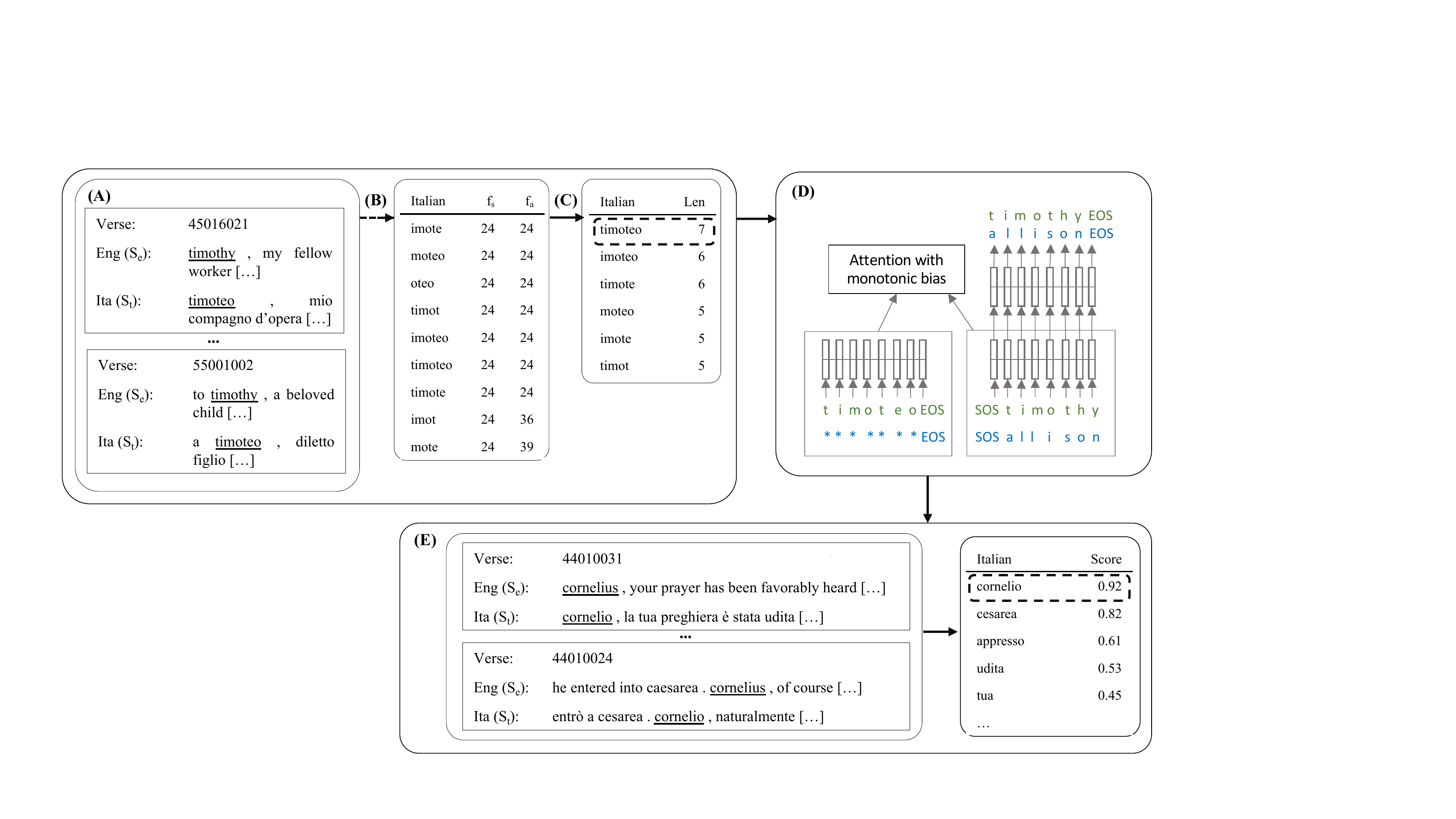} \vspace*{-4mm} \caption{Data
		flow in \methodn. Example showing
	extraction of
		Italian NE training candidates for English
	``timothy'' and identification of an Italian NE that matches English ``cornelius''.
		The input is the parallel corpus (A). 
		\methodb extracts from the parallel corpus  ngrams that are candidate transliterations for ``timothy'' (B).  
		These candidates are then filtered (C). 
		(D): The architecture of the neural transliteration model. 
		Green input-output pairs: Italian-English training data
		taken from the output of CLC-B.
		Blue input-output pairs: monolingual English training data.
		(E): We use  the trained neural model
to score candidates 
taken from the Italian parallel verses in which ``cornelius'' appears and keep the best scoring word.
		\label{fig:schema}}
\end{figure*}

\section{Related work}

\subsection{Word alignment} 
A multilingual named entity resource can be extracted from a
parallel corpus via word alignment. 
Word alignment  has been widely studied. Statistical word alignment models were introduced by \newcite{brown1993mathematics}. More recently Giza++ \cite{och2000improved} and Eflomal \cite{ostling2016efficient} were released followed by neural network extensions \cite{ngo2019neural}. Other approaches use learned representations for creating alignments \cite{jalili-sabet-etal-2020-simalign}. 
In concurrent work, \newcite{imani2021graph} have shown that better word alignment
results can be achieved by exploiting multiparallel corpora.
Previous work on named entity alignment and recognition uses combinations of alignment tools and postprocessing techniques. \newcite{dasigi2011named} use Giza++ for alignment   and applied statistical machine translation \cite{koehn2007moses} and language-specific rules for improving transliteration. \citelanguageresource{wu2018creating} use the Berkeley aligner \cite{liang2006alignment} to word-align language pairs in the English Bible and further improve them with machine translation. 
In this paper, we do not use word aligners because of their
low quality for named entities in small parallel corpora. We
will directly compare with the word-alignment-based method
of 
\citelanguageresource{wu2018creating}.

Recent approaches rely on parallel corpora and multilingual
pre-trained models.  \newcite{wu2021unitrans} construct a
pseudo training set by performing translation and use
multilingual BERT \cite{devlin2018bert} to generate language
independent features for training NER
models. \newcite{li2021cross} use
XLM-R \cite{lample2019cross} to build an entity alignment
model that projects English named entities into the
parallel target language. While these approaches are
promising, they are limited to the language set the models
have been trained on ($\approx$100). In contrast we
apply \methodn to the more than one thousand languages in
the Parallel Bible Corpus.

\subsection{Transliteration}
\newcite{prabhakar2018machine} provide a
comprehensive survey on transliteration.
Recently, the task has been addressed with sequence-to-sequence
models and
transformers. 
\newcite{wu2018comparative}
perform
experiments with these models on 
their Bible-based translation matrix dataset \citelanguageresource{wu2018creating}
and show that
the task is challenging in the low-resource scenario. One of
the causes is overfitting of the training set due to its
reduced size. Our \methodn method uses a transliteration model and
addresses this problem by augmenting the training set with
monolingual target data (English) and introducing a
monotonic bias.

\subsection{Named entity resources}

\citelanguageresource{benites-etal-2020-translit}
introduce Translit, a transliteration resource created by
combining and unifying public corpora. However, this dataset
only covers 180
languages. BabelNet \citelanguageresource{navigli2012babelnet} is a
multilingual encyclopedic dictionary that integrates
WordNet, Wikipedia, GeoNames, inter alia. BabelNet is more
comprehensive than other resources, but its NE coverage is
still poor for many languages (e.g., for Inuktitut).
We show in this paper that we can extend the coverage of
BabelNet with our method.
The Translation Matrix
of \citelanguageresource{wu2018creating}  covers 591 languages. Their
approach is based on word alignment.  We 
show that our approach  yields higher quality.

\subsubsection{Named Entity Recognition resources}
Named Entity Recognition (NER) systems usually require annotated data to achieve high accuracy. Our NE resource can be exploited to 
bootstrap
such NER models for many different languages.
\citelanguageresource{al2015polyglot}  automatically extract named entities from Wikipedia link structure and Freebase attributes and create Polyglot-NER for 40 languages.
\citelanguageresource{pan2017cross} introduce WikiAnn,
a resource for 282 Wikipedia languages that supports name
tagging and entity linking.
Our resource covers more than 1300 languages and \methodn does not rely on external sources other than the PBC.

\subsection{Annotation projection}
\citelanguageresource{ehrmann2011building} project annotations from English to five languages using a 
 phrase-based statistical machine translation system and different methods: string matching, consonant signature matching and edit distance similarity.
\newcite{ni2017weakly} propose two methods for NER projection using heuristics, alignment information, and mapped word embeddings. 
\newcite{wang2018cross} describe a method for cross-lingual knowledge graph alignment of pre-aligned entities based on their distance in the learned embedding space.
 We project English NEs to the target languages exploiting character-level correspondence and a neural transliteration model without requiring any word alignment information or seed data.

\subsection{Monotonicity}
The performance of sequence-to-sequence
models on some tasks can be improved by imposing an
inductive bias of monotonicity (i.e., no character can be aligned to one that precedes a previously aligned character).
Previous studies implement and analyze the effect of such a monotonic bias.
\newcite{wu2019exact} show that enforcing strict
monotonicity and learning a latent alignment jointly while
learning to transduce leads to improved performance for
morphological inflection, transliteration, and
grapheme-to-phoneme conversion. \newcite{rios2021biasing}
develop a general method for incorporating monotonicity into
attention  for seq2seq and Transformer models, agnostic of the task and model architectures.
Similar to this prior work, we impose a monotonic bias on our neural transliteration model.

\section{Method}
We now describe \methodn.\footnote{Reproducibility details
in \secref{repr}.}
Figure \ref{fig:schema} shows architecture and data flow.
For ease of development and evaluation, we also use the
Uroman romanizer \citelanguageresource{hermjakob2018out}. It
converts scripts into Latin characters. But \methodn can be applied equally well without romanization. 
\methodn consists of two steps. First we extract
character-level correspondences (\methodb). Then we train a neural transliteration model to obtain the final set of named entities.

\subsection{Character-Level Correspondence Bootstrapping (CLC-B)}
\label{secboot}
We use cooccurrence statistics at the character level between English NEs and  target language NEs to create a training set for the neural transliteration model.
We use \citelanguageresource{wu2018creating}'s list of
English Bible NEs. 
NEs with frequency 1 are not considered in \methodb because the FILTER step (\#3 below) is likely to produce false positives (accidentally correlated ngrams) for them; but they are considered in \S\ref{sectranslit}.

\methodb is designed based on the following simple correspondence
assumption:
if an English NE occurs in a verse, the corresponding
target NE occurs in the parallel target verse
and vice versa. This also implies that 
if  $N$ and $M$ are the 
the number of verses in which the NE and its translation
occur, then
$N \approx M$. We do not require $N=M$ because we relax the
correspondence assumption due to errors in the parallel
corpus and due to the use of pronouns (including null
pronouns, i.e., the pronoun is only present implicitly), which differs across languages.

We now describe our Character-Level Correspondence Bootstrapping
(\methodb) method, for the example of an English NE $w$. Algorithm \ref{fig:algo} shows the pseudocode. 
Let  $f_a$ be the total frequency of an ngram in the target
language and $f_s$ its frequency in the subset of verses that contains
$w$ in English.

\begin{enumerate}
    \item 
    \textbf{EXTRACT.} (Line 4)
    Extract the parallel subcorpus that contains $w$ from the parallel corpus. It consists of the English part $S_e$ and the target language part $S_t$.
    \item 
    \textbf{GET\_NGRAMS.} (Lines 5--13)
    For all character $n$-grams\footnote{We discard ngrams containing digits, punctuation 
    and spaces.} 
    ($3<n<20$) in $S_t$, determine $f_s$, the number of occurrences in $S_t$. 
    Discard ngrams with $f_a > 50$ -- this removes a  small number of frequent NEs like Jesus, but avoids false positive matches with frequent ngrams. The resulting set of target ngrams is $G_t$.
    \item 
    \textbf{FILTER.} (Line 14)
    Filter $G_t$ as follows.
    \begin{enumerate*}
        \item Determine the ngram(s) with the highest $f_s$. Remove all other ngrams.
        
        \item Determine the ngram(s) with the minimum absolute difference between
        $f_a$ and $f_s$. Remove all other ngrams. Intuitively, most NEs in a particular domain are unique -- so they should contain ngrams that only occur in the NE and not in other words.
        
        \item Return the ngrams with the smallest length difference to $w$. This
        eliminates candidates that are much longer or shorter than $w$.
    \end{enumerate*}
\end{enumerate}

	\begin{algorithm} 
		\small
		\algrenewcommand\algorithmicindent{0.2cm}
		\def\tl#1{\text{#1}}
		\begin{algorithmic}[1]
			\Procedure{\methodb}{corpus $E$, corpus $T$, list $\tl{English\_NEs}$} 
			\State $pairs \leftarrow list()$
			\For {$\tl{w} \in \tl{English\_NEs}$}
			\State $S_e,S_t \leftarrow \tl{extract}(\tl{w}, S, T)$ 	\Comment{(1) EXTRACT}
			\State $G_t \leftarrow list()$
			\State $\tl{ngram\_list }\leftarrow \tl{get\_ngram\_list}(S_t)$
			\State $\tl{frequency\_list} \leftarrow \tl{get\_frequent\_ngrams}(S_t)$
			\For {$[\tl{ngram}, \tl{count}] \in \tl{ngram\_list}$}
			\If{ $\tl{ngram} \in \tl{frequency\_list}$ or $\tl{count}==1$}
			\State continue
			\EndIf
			\State $G_t.append([\tl{ngram},\tl{count}])$
			\EndFor
			\State $\tl{pairs}.append(\tl{filter}(G_t))$	\Comment{(3) FILTER}
			\EndFor
			\State \Return $\tl{pairs}$
			\EndProcedure
		\end{algorithmic}
		
	\caption{Pseudocode for the \methodb method.
        Given a parallel corpus of English ($E$) and a
		target language ($T$), we identify, for each
		English NE, its target match.
                See \S\ref{secboot} for details and for the
        EXTRACT and FILTER methods.
	\label{fig:algo}}
	\end{algorithm}
	

\subsection{Neural transliteration}
\label{sectranslit}
\methodb returns a noisy set of NE pairs, especially when
        only a small number of parallel verses is available
        for a language (we refer to this as the
        \emph{lowest-resource} setting below). We build a neural sequence-to-sequence model \cite{sutskever2014sequence} to refine it and to mine additional pairs.
We use a single-layer bidirectional Gated Recurrent Unit (GRU) \cite{cho-etal-2014-learning} encoder and a single-layer GRU decoder with attention \cite{luong2015effective}. 
The sequences are processed at the character-level, with separate input and output vocabularies.
Target language NEs are the input, English NEs the output; we use input/output when referring to the neural model (not source/target) because ``target'' in this paper refers to the target language that English is paired with.

To make best use of the limited training data in our experimental
setup, we use augmentation and impose a monotonicity bias as
described below.
To avoid overfitting, we augment the training set with English NEs. We label the English Wikipedia dump\footnote{\url{https://dumps.wikimedia.org/} (01.04.2020)}
with the Flair Part-of-Speech tagger \cite{akbik2019flair}, and select all NEs.
We add,
for each English NE mined from Wikipedia,
one pair of the form (empty input NE, English output NE) to the training set.
We use empty input NEs to prevent the learning of the identity function while helping the decoder to learn the structure of English words. 
To prevent generation of output independent of the input, we ensure equal proportions of original and augmented data by oversampling the former.
Because transliterations are (with few exceptions) monotonic, we impose a monotonicity bias: we mask the attention matrix, so that the model cannot
see anything to the left of the position previously attended
to.

Given an English NE $w$ and the verses $S_e$ in which it appears,
target candidates are all words in $S_t$, the verses
parallel to $S_e$. Once the model is trained, we choose the best scoring candidate as $w$'s transliteration where the score is the average
log likelihood of the output characters \cite{severini-etal-2020-combining}.

We use a slightly different scoring step for non-tokenized languages (e.g., Japanese) because separated words in $S_t$ are not available: given an English NE $w$, the target candidates are all ngrams that \methodb has extracted for $w$ in step 3b. 

\begin{table}[!t]
	\centering
	\small
	\resizebox{.4\textwidth}{!}{%
		\begin{tabular}{lllrr}
			\hline
			&Lang    & ISO &  \# verses    & \# parallel  \\ \hline
			\multirow{7}{*}{\rotatebox{90}{\scriptsize\begin{tabular}{c}low-resource\\ languages\end{tabular}}}
			&Arabic  & Arb &  31173        &  31062 \\
			&Finnish & Fin &  31167        &  31061 \\
			&Greek   & Ell &  31183        &  31062 \\
			&Russian & Rus &  31173        &  31062 \\
			&Spanish & Spa &  31167        &  31062 \\
			&Swedish & Swe &  31167        &  31062 \\
			&Zulu    & Zul &  31167        &  31062 \\ \hline
			\multirow{6}{*}{\rotatebox{90}{\scriptsize\begin{tabular}{c}lowest-resource\\ languages\end{tabular}}}
			&Hebrew  & Heb &  7952         &  7917  \\  
			&Hindi   & Hin &  7952         &  7917   \\
			&Kannada & Kan &  7952         &  7917  \\
			&Korean  & Kor &  7913         &  7869  \\
			&Georgian & Kat & 4904   &  4844 \\
			&Tamil   & Tam &  7942         &  7917  
		\end{tabular}
	}
	\caption{Number of verses in PBC and
		number of verses that are parallel with our English edition
		for the languages in our experiments.
		The English edition has 31,133 verses. 
		\label{tab:stats}}
\end{table}

\def\maintablesep{0.1cm}

\begin{table*}[!ht]
	\scriptsize
	\renewcommand{\arraystretch}{1.1}
	\begin{tabular}{l|r@{\hspace{\maintablesep}}r|r@{\hspace{\maintablesep}}r|r@{\hspace{\maintablesep}}r|r@{\hspace{\maintablesep}}r|r@{\hspace{\maintablesep}}r|r@{\hspace{\maintablesep}}r|r@{\hspace{\maintablesep}}r||r@{\hspace{\maintablesep}}r}
		& \multicolumn{2}{c|}{Arb} & \multicolumn{2}{c|}{Ell} &  \multicolumn{2}{c|}{Fin}  &
		\multicolumn{2}{c|}{Spa} & \multicolumn{2}{c|}{Swe} & \multicolumn{2}{c|}{Rus}  & \multicolumn{2}{c||}{Zul}  & \multicolumn{2}{c}{AVG}  \\
		
		& Dist  & Hum  & Dist & Hum  & Dist   & Hum  & Dist  & Hum & Dist  & Hum & Dist   & Hum  & Dist   & Hum & Dist   & Hum\\ \hline
		
        \citelanguageresource{wu2018creating}
		& 67.9   & 70.0  
		& 47.2   &   80.0 
		& \underline{89.0}   &  90.0 
		& 87,6   &   91.7
		& 88.8   &  88.3 
		& 60.6   &   72.9  
		& 61.9   &  84.8 
		
		& 65.9   &  82.5 
		\\
		
		\newcite{ostling2016efficient}
		& 69.8  &  61.7 
		& 53.4  &  88.3 
		& 77.7  &  76.7
		& 83.9  & 86.7 
		& 81.2  & 85.0 
		& 64.8  & 83.1
		& 52.9  & 86.4 
		
		& 60.9 &  81.1 \\

		\newcite{sabet2020simalign}      
		& 18.1    & 20.0  
		& 23.5  &   40.0 
		& 49.8   & 60.0 
		& 35.6   &  45.0 
		& 41.6   & 50.0 
		& 39.6   & 45.8 
		& 18.3  & 25.4 
		
		& 29.6  &  40.9  \\
		
		\methodb
		& 53.8   & 56.7
		& 32.2   &  45.0 
		& 59.9   & 50.0 
		& 48.0   &  48.3
		& 52.0   &  48.3 
		& 46.5   &  57.6
		& 55.1   &   74.6 
		
		& 46.6  & 54.4  \\ 
		
		\methodn   
		& \underline{70.6}   &  \underline{81.7}   
		& \underline{54.7}   & \underline{91.7} 
		& 86.5  &  \underline{93.3} 
		& \underline{89.6}   & \underline{96.7}
		&  \underline{89.9}   & \underline{91.7} 
		& \underline{70.2}   & \underline{84.8} 
		& \underline{68.8}     & \underline{93.2} 
		
		& \underline{71.9} &  \underline{90,4}
		
	\end{tabular}
	\newline
	\vspace*{0.05 cm}
	\scriptsize
	
	\setlength{\tabcolsep}{5.5pt}
	\begin{tabular}{l|r@{\hspace{\maintablesep}}r|r@{\hspace{\maintablesep}}r|r@{\hspace{\maintablesep}}r|r@{\hspace{\maintablesep}}r|r@{\hspace{\maintablesep}}r|r@{\hspace{\maintablesep}}r||r@{\hspace{\maintablesep}}r}
		\multicolumn{1}{c|}{}  
		& \multicolumn{2}{c|}{Heb}    & \multicolumn{2}{c|}{Hin}      & \multicolumn{2}{c|}{Kan}    & \multicolumn{2}{c|}{Kat} & \multicolumn{2}{c|}{Kor} & \multicolumn{2}{c||}{Tam}  & \multicolumn{2}{c}{AVG}   \\

		\multicolumn{1}{c|}{}    & Dist  & Hum  & Dist & Hum  & Dist   & Hum  & Dist  & Hum & Dist  & Hum & Dist   & Hum &Dist  & Hum   \\ \hline
		
		\citelanguageresource{wu2018creating}
		& 53.4    & 62.5
		& 64.1$*$ & 76.3$*$
		& 41.5    &  61.7 
		& 64.3    &    70.0 
		& 30.5    &  54.2 
		& 47.1$*$ &  66.1$*$
		
		& 50.2    &   65.1 \\
		
		\newcite{ostling2016efficient}
		&  \underline{65.5}  & \underline{83.9}
		& 57.7$*$  & 69.5$*$
		& 23.1 & 38.3 
		& 64.0  & 68.3 
		& 16.6  & 33.9
		& 20.5$*$  & 35.6$*$
		
		& 41.2  & 57.5  \\

		\newcite{sabet2020simalign}   
		& 27.8      &  23.2 
		& 41.6$*$    & 47.5$*$
		& 26.5       & 46.7 
		& 28.0       & 20.0
		& 23.5       & 40.0
		& 30.4$*$    & 47.5$*$
		
		& 29.6      & 37.5
		\\ 
		
		\methodb
		& 37.2     &  51.8 
		& 43.1$*$  &  39.0$*$
		& 30.7    &  48.3 
		& 41.7    & 45.0 
		& 23.8     &  37.3 
		& 38.6$*$  &   47.5$*$
		
		& 35.9    & 44.8
		\\
		
		\methodn   
		& 62.6     & 	71.4	
		& \underline{78.6}$*$  &  \underline{94.9}$*$
		& \underline{45.9}     & \underline{93.3}	
		& \underline{70.8}     &   \underline{88.3} 
		& \underline{34.2}     & \underline{78.0}	
		& \underline{59.5}$*$  &   \underline{91.5}$*$
		
		& \underline{58.6}     & \underline{86.2}
		
	\end{tabular}
	\caption{Precision of NE correspondence identification for
		low-resource (top: Hebrew Bible and New Testament) and
		lowest-resource (bottom: New Testament only) languages. We
		compare Translation Matrix \protect\citelanguageresource{wu2018creating}, Eflomal  \protect\cite{ostling2016efficient},
		SimAlign \protect\cite{sabet2020simalign},
\methodb and \methodn. Comparisons are with silver data+Jaro
	distance (Dist) and with gold human annotated data
	(Hum). *: evaluation on romanization for fair comparison with baselines.  \label{tab:mainresults}}
\end{table*}

\section{Evaluation and Analysis}\label{sec:evaluation}
We apply \methodn to the Parallel Bible Corpus
(PBC) \citelanguageresource{mayercreating2014} for evaluation and for
creating our NE resource.\footnote{Reproducibility details
	in \secref{repr}}
We evaluate on a subset of 13
languages that includes different scripts, resource
availabilities and language families: Arabic, Greek,
Finnish, Hebrew, Hindi, Kannada, Korean, Georgian,
Russian, Spanish, Swedish, Tamil, and Zulu.  These languages
are also covered by the baselines and are therefore suitable
for comparison.  We view them as a representative subset
for evaluating our method's performance.
Note, however, that our NE resource covers all 1340 PBC languages:
our approach is applicable to all languages since it does not use
language-specific features and preprocessing steps.

PBC contains 1340 languages, most of which are
low-resource. It is divided into subfiles, each containing
Bible text from one language.  Some languages that cover the
Hebrew Bible and the New Testament completely contain about
30,000 verses. Other languages
contain fewer than 8000
verses.   We divide the languages into two
categories:
\textbf{lowest-resource}, fewer than 8000  verses; and
\textbf{low-resource}, between 8000 and 32,000
verses.\footnote{``low-resource'' is to be interpreted as
referring to the setting in our experiments. For example,
many resources are available for Russian, but in our setting
we only use the Russian text that is available in PBC to evaluate how well
our method works in a low-resource setting.}
\tabref{stats} gives the number of verses for the
editions we use.
We evaluate our resource on human annotated data and on silver data with respect to the baselines and provide analysis.

\subsection{Human evaluation}\label{sec:gold}
We annotated 60 NEs per language using Toloka,\footnote{\url{https://toloka.yandex.com/}} a crowd-sourcing platform.
Annotators had to pass an English test and successfully complete a training task to gain access to the annotation pool. Their performance was constantly checked using covert control questions.
Each question contained the English NE and up to five
possible options: one for each of the three baselines, one for \methodb 
and one for \methodn. Each option consists of the word in the target script together with its romanized version in parentheses. Annotators had to mark all correct options that can be paired to the English NE, or none if no option is correct. Each question was annotated by exactly three annotators.

We calculate annotator agreement using  Cohen's Kappa  \cite{cohen1960coefficient}, which measures  agreement above chance.
Similar to the setup of \citelanguageresource{wu2018creating}, we do not
require that the annotators know the target
languages. However, their average pairwise agreement is
0.73, ``substantial agreement''
according to Cohen's Kappa  \cite{landis1977measurement}, indicating that they
can find the correct corresponding target named entity even
if they do not know the target language.
To create the final gold set, we adopt a majority voting
strategy and keep named entities
that at least two annotators agreed on, resulting in at
least 58 named entities per language.

We evaluate \methodn  and the baselines on this gold set.\footnote{We release the gold dataset to facilitate future research.} The results can be found in \tabref{mainresults}, column ``Hum''. 
\methodn outperforms the baseline \citelanguageresource{wu2018creating} for all languages (average
difference of 7.9), with substantial improvements for the
lowest-resource languages (difference of 21.1).
The biggest improvements are for Hindi and Kannada (more than 30).

\subsection{Silver evaluation}
The gold dataset is used as the main evaluation of the
resource.
However, we additionally create a silver
dataset to 
evaluate based on a larger set of hundreds of NEs.
We create the silver set
by translating each English NE to all target languages
supported by the Google translation
API\footnote{\url{https://cloud.google.com/translate}} and
comparing them with the NEs extracted by \methodn using the
Jaro distance \cite{jaro1989advances}. The distance takes
into account the number and order of characters shared by
two strings; e.g., the NE ``salome'' has a distance of
$0.05$ from ``salom'' and $0.11$ from ``calom''. Jaro is
frequently used for entity matching and is well-suited for
short strings \cite{cohen2003comparison}.  We use a
threshold of $0.3$ for the Jaro distance, chosen to be
strict enough to evaluate the NEs and to take into account
noise in the pairs produced by Google Translate.  For
example, the silver translation of "jannes" in Greek is
\begin{otherlanguage*}{greek}γιάννες\end{otherlanguage*}  (giánnes) while our data contains
\begin{otherlanguage*}{greek}ιαννής\end{otherlanguage*}
(iannís), which  is also correct; their distance is $0.26$. 
Another example is the name "mitylene" that the silver data translates to
\begin{otherlanguage*}{greek}μυτυλένιο\end{otherlanguage*}  (mytylénio) and has a distance of $0.27$ to our translation  \begin{otherlanguage*}{greek}μυτιλήνη\end{otherlanguage*} (mytilíni).

By design, the silver data provides only a single translation for each English NE.
However, multiple translations are often correct, due to the variability of morphology, transliteration, naming conventions and dialects
\cite{prabhakar2018machine}. For example, the English NE ``Paul'' can be aligned to  Russian ``Pavel'' and ``Pavla''.  For this reason,
our results on the silver standard must be interpreted as lower bounds.

Arabic and Hebrew are standardly written without short vowels. This is also the case for the silver data. However, some PBC editions are written with short vowels, so we postprocess predictions by removing short vowel diacritics.

\tabref{mainresults} shows results for the 13
languages.
The ranking of baselines and methods is similar to the one obtained with the gold human evaluation with \methodn being always the best, except for Finnish.
Improvements for lowest-resource languages (lower part of the table) are large,
up to 48\% difference on average. 
\methodn outperforms \citelanguageresource{wu2018creating} for 12 of the 13  languages.\footnote{
The exception is Finnish, which is probably due to the fact
that machine translation (which was used for \citelanguageresource{wu2018creating}) performs
well for high-resource languages. Note however,
that \methodn performs best for Finnish in the
(more reliable) human (``Hum'') evaluation.}

\subsection{Word alignment comparison}
NE correspondences can also be obtained using a word aligner. 
We compare our results with pairs obtained using Eflomal \cite{ostling2016efficient}, a statistical word aligner,  and SimAlign \cite{sabet2020simalign}, a high-quality word aligner that leverages multilingual word embeddings.
\tabref{mainresults} shows precision for silver and gold data.
\methodn outperforms Eflomal (with the exception of Hebrew) and SimAlign for all 13 languages. We attribute this to the fact that NEs are hard to word-align because most of them are infrequent, resulting in alignment errors due to sparseness.
\methodn could be integrated into word alignment pipelines to boost word aligner performance for NEs \cite{sajjad2011algorithm,semmar-saadane-2013-using}. 

\methodb works at the character level and is applicable to non-tokenized languages 
while aligners are not.
Japanese is non-tokenized, so we evaluate it (only
for \methodb and \methodn since the other methods were not
run on Japanese).
We evaluate the 979 pairs of \methodn with the silver data and obtain a precision of 63.2\%.
We also use Toloka for the gold evaluation of 60 random
pairs and obtain a precision of 60\%. However, in this case
each question has at most two options (\methodb and \methodn\
-- in contrast to five as for \ref{sec:gold}), which can hinder the annotators' judgments having less comparison terms. 
For this reason, we also asked three experts to evaluate the 60 pairs and obtained a precision of 85\%. 

\begin{table*}[!t]
	\small
	\centering
	\begin{tabular}{r|llllllll} \hline
		\# & English   & Arabic        & Finnish  & Greek    & Hebrew   & Kannada      & Russian   & Tamil    \\ \hline
		28 & elijah  &	alalihaau & eliaa & elia & 	veaeliyahu & eliiyanaagali & elisei & eliyaavaa \\
		12 & titus & tiytusa & titus & titos & titos & titanannu & titu & tiittuvin \\
		8  & elizabeth & aaliysaabaata & elisabet & elisabet & elisheva & elisabeet    & elizaveta & elicapet \\
		3 & miletus & miyliytusa & miletokseen & mileto & lemilitos & mileetakke &mileta & mileettu \\
		2  & rufus     & ruwfusa       & rufuksen  & roufo   & vishelom  & uphaniguu
		& rufa      & ruupuvukkum     \\
		2  & hermes    & wahirmisa       & hermeeksi   & epairne  & heremes  & meeyaniguu
		& germes    & ermee    
	\end{tabular}
	\caption{Examples of named entity alignments (romanized). ``\#'' column shows the number of verses in which the  English word appears.	\label{tab:examples}}
\end{table*}

\subsection{Impact of corpus size}
\tabref{mainresults} shows that precision for
lowest-resource languages (less than 8000 verses, bottom) is worse than
those for low-resource languages (about 30,000 verses, top), with an
average difference of $13.3\%$ for silver data, and
$4.2\%$ for gold data.
The small gap on gold data, 
  highlights that our method is 
  appropriate also for the lowest-resource setting.
Table \ref{tab:examples} shows some examples of aligned
pairs according to \methodn.
We see that  errors
arise
as the frequency of NEs in
the English corpus diminishes.
For example,
the Kannada alignment for ``rufus'' and
Greek and  Kannada
alignments for ``hermes'' are incorrect. Both words are short,
indicating another source of errors: short words provide
less of a signal for the neural transliteration model  than long words do. 

\subsection{Impact of neural transliteration}
Table \ref{tab:mainresults} shows precision
for \methodb and \methodn.
All languages benefit from neural transliteration
with an average improvement of 30.9 percentage points.
One of the reasons is that \methodb was designed to discard English NEs that appear only once in the corpus. 
Table \ref{tab:freq_vs_neural} shows examples where neural transliteration corrects an error made by \methodb. Most of these cases have low frequency. This is not surprising as the risk of false positives increases as the frequency decreases because the heuristics used in \methodb (\S\ref{secboot}) are less reliable for low-frequency NEs.

\begin{table}[!t]
\small
\centering
\setlength{\tabcolsep}{3.8pt}
\begin{tabular}{lllll} \hline
Lang & Eng & Freq 
& \methodb        & \methodn   \\ \hline
Arb & anah
& 10
& \includegraphics[width=0.03\textwidth, height=0.02\textwidth, trim={0 7.7cm 5cm 0}]{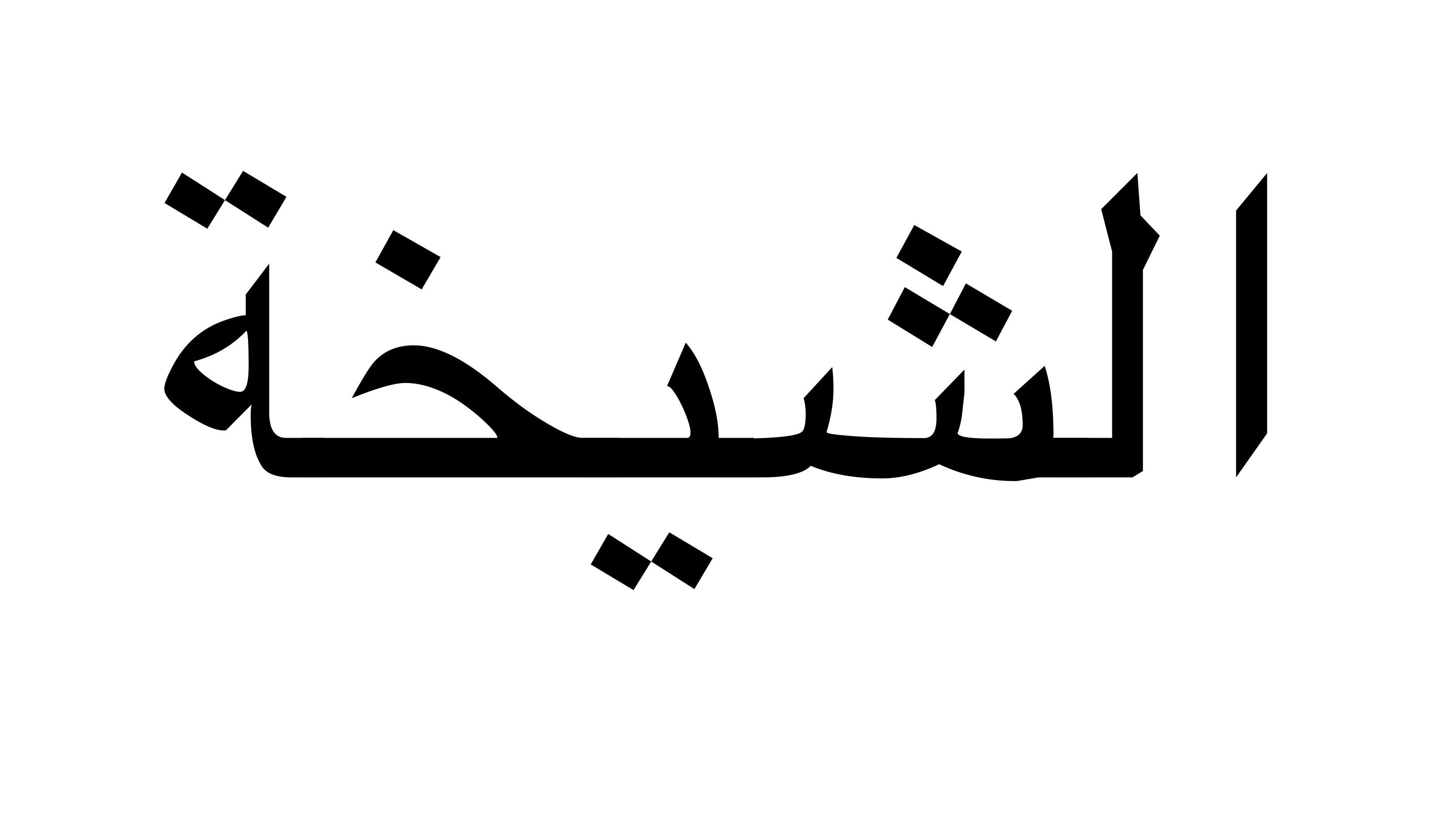} (alshaykha)

& \includegraphics[width=0.015\textwidth, height=0.02\textwidth, trim={12cm 7.7cm 10cm 0}]{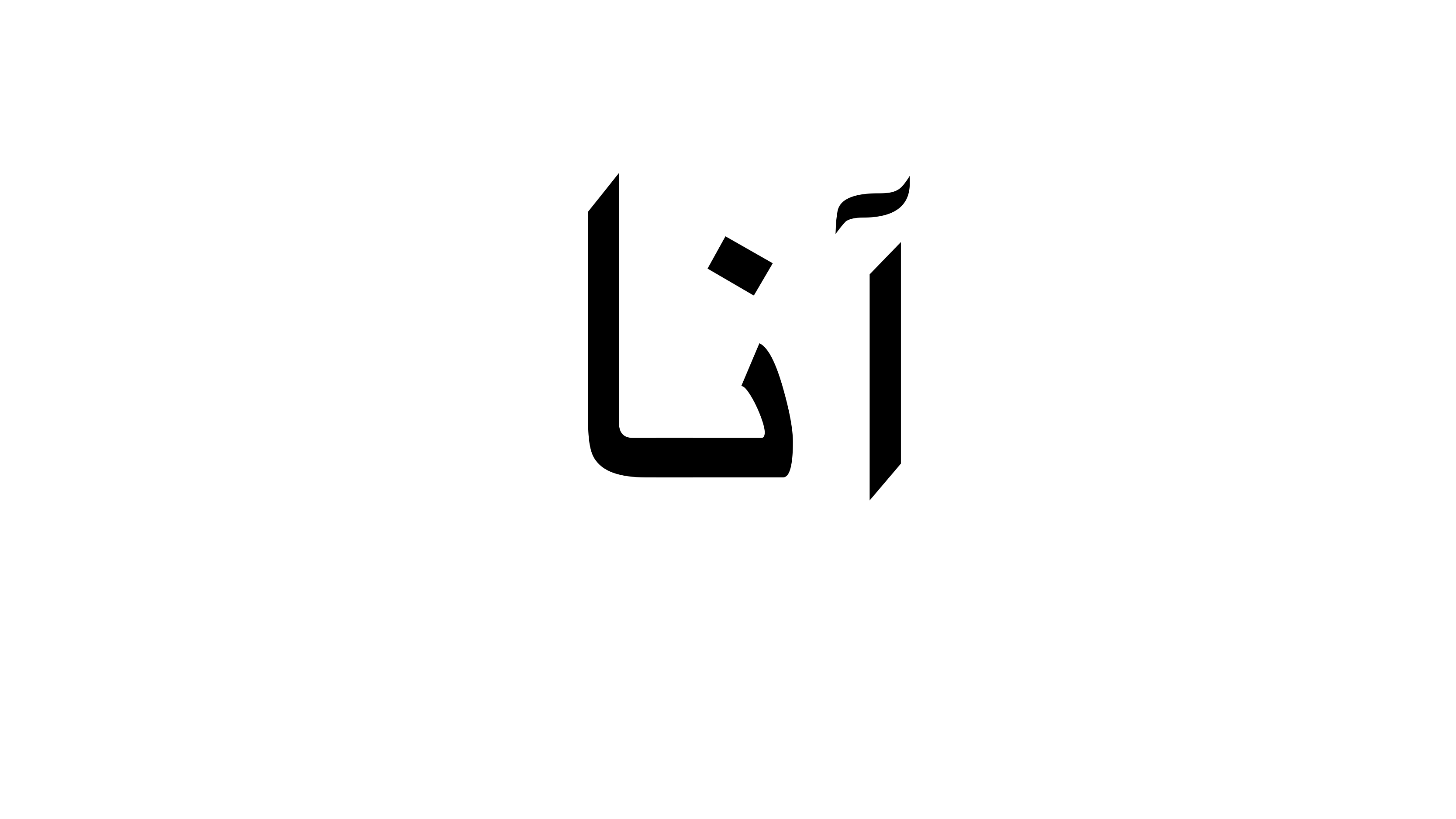} (ana)

\\
Rus  &  joanna
& 2
&\includegraphics[width=0.045\textwidth, height=0.013\textwidth, trim={10.5cm 8.6cm 10.5cm 6.9cm}, clip]{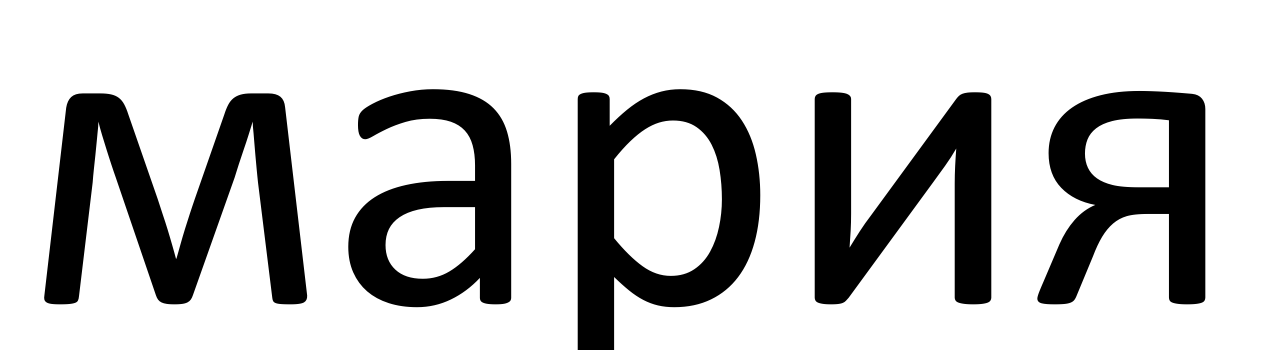} (mariya)          &   \includegraphics[width=0.05\textwidth, height=0.012\textwidth, trim={10cm 9cm 10cm 7.1cm}, clip]{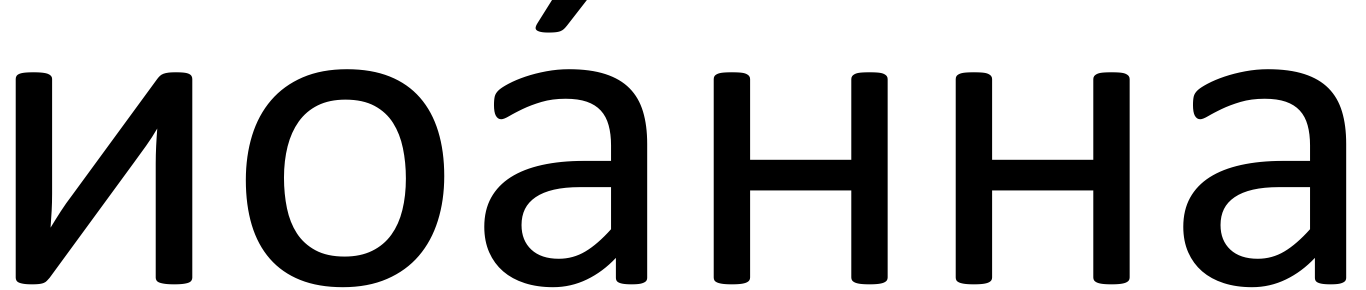}  (joanna)      \\

Fin  & perez  
& 2 
& hesroni        & peresin  \\

Kan & cainan & 2 &   \includegraphics[width=0.03\textwidth, height=0.02\textwidth, trim={5cm 7.7cm 5cm 0}]{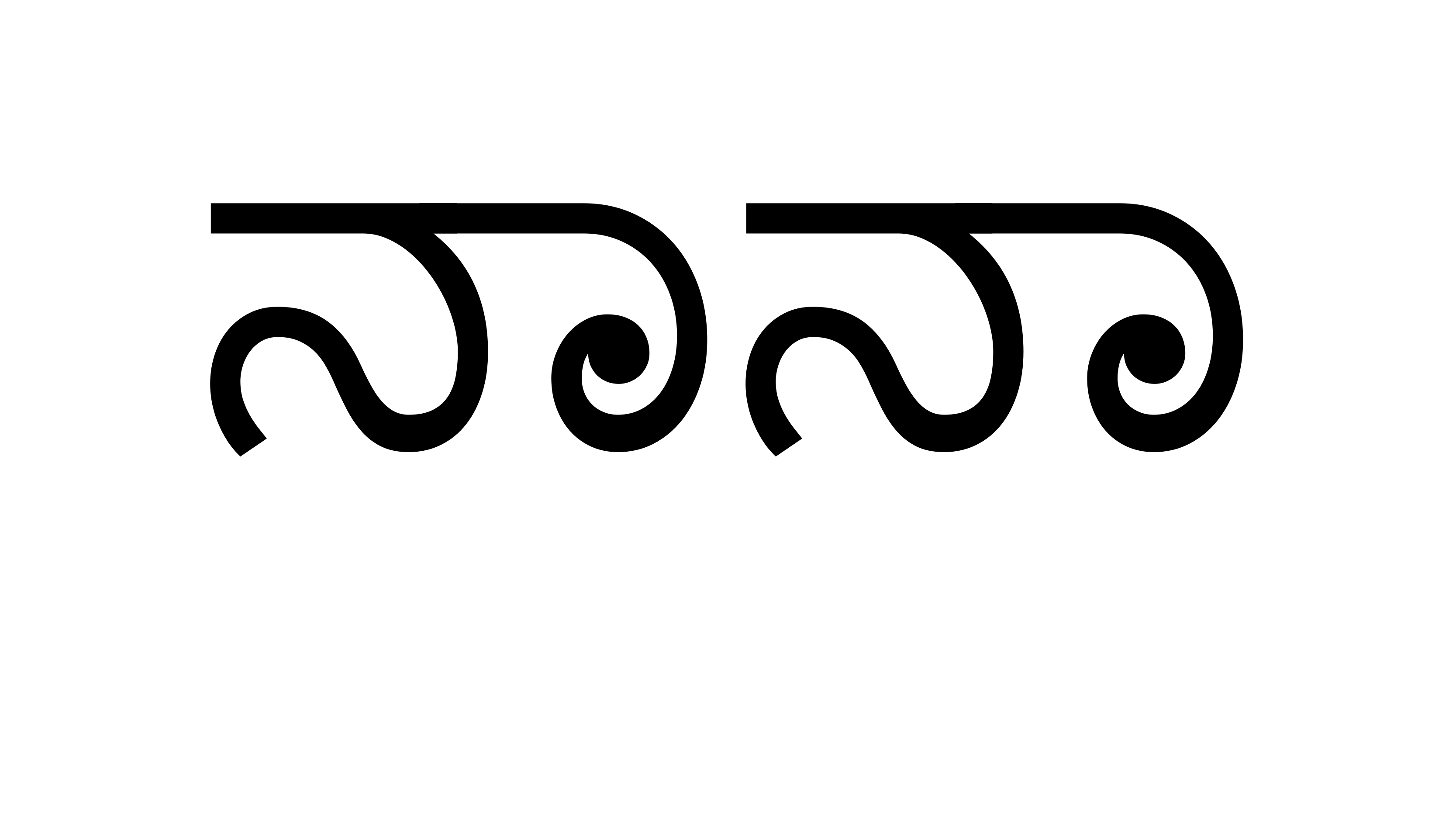} 

&  \includegraphics[width=0.06\textwidth, height=0.02\textwidth, trim={3cm 9.7cm 4cm 0}]{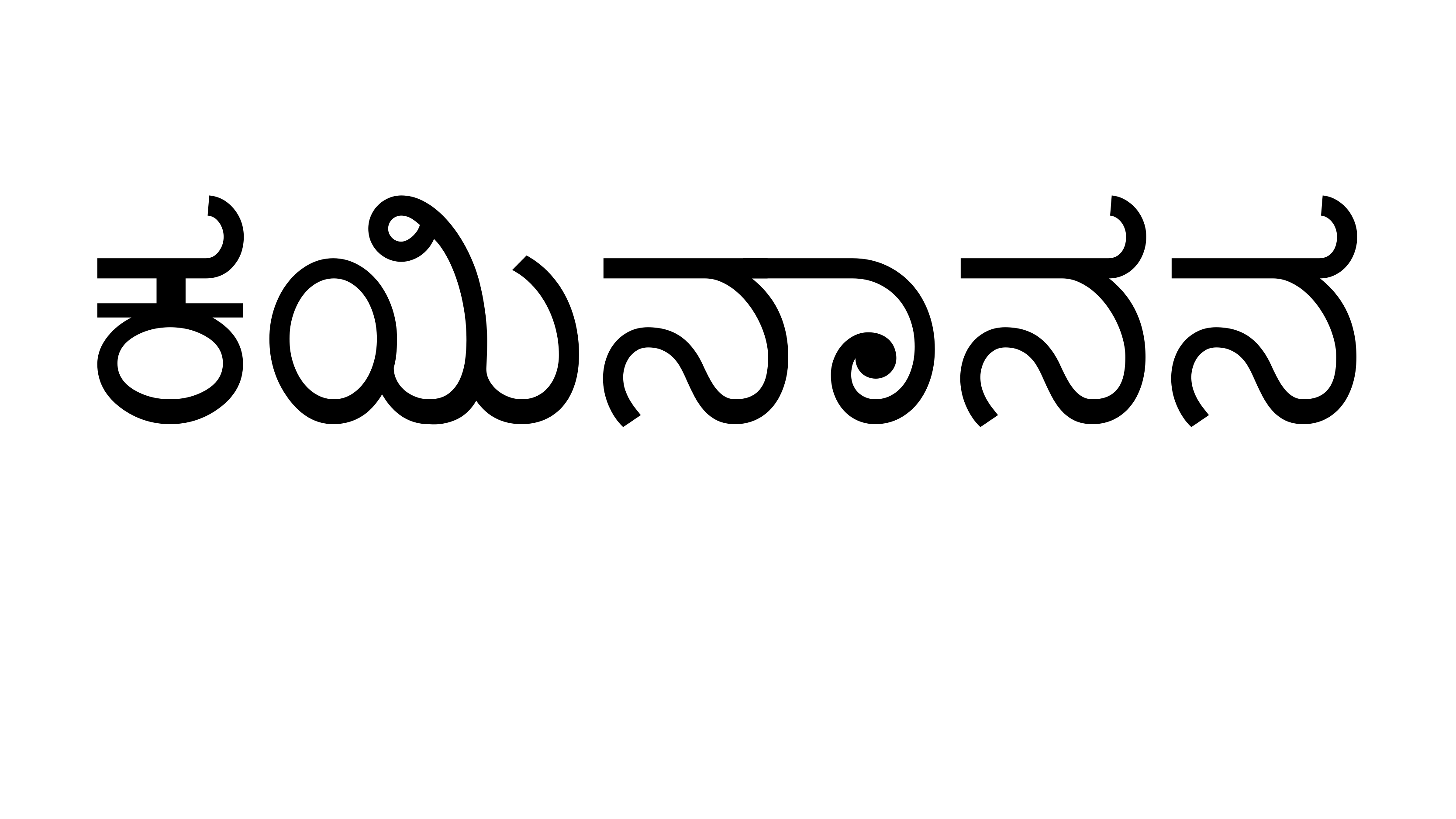}  \\
&&&(naanaa)& (kayinaanana) \\

Tam  & azor   
& 2 
&	 \includegraphics[width=0.07\textwidth, height=0.022\textwidth, trim={2cm 9.7cm 4cm 1cm}]{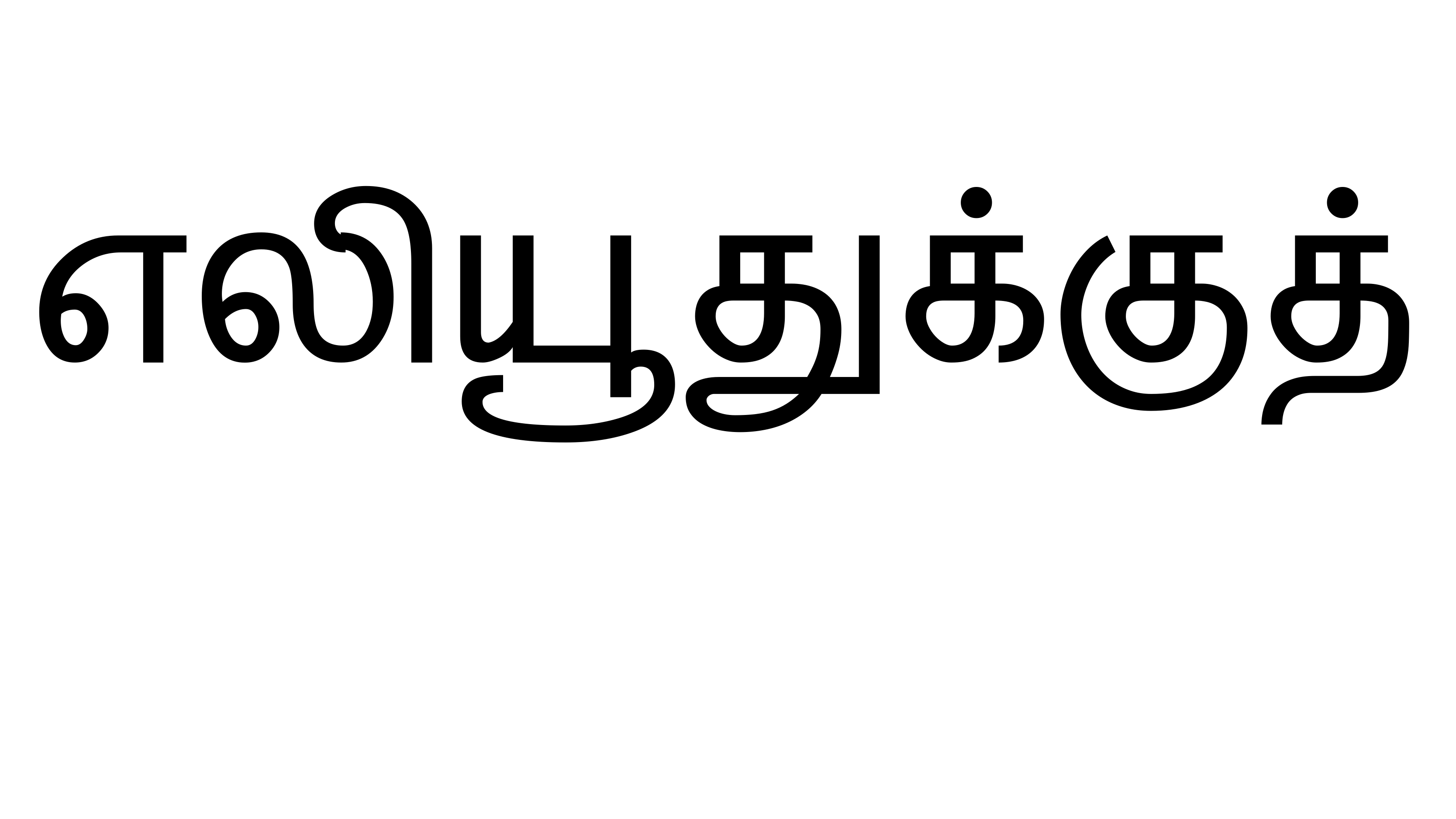}   

& 	 \includegraphics[width=0.07\textwidth, height=0.022\textwidth, trim={7cm 9.7cm 4cm 1cm}]{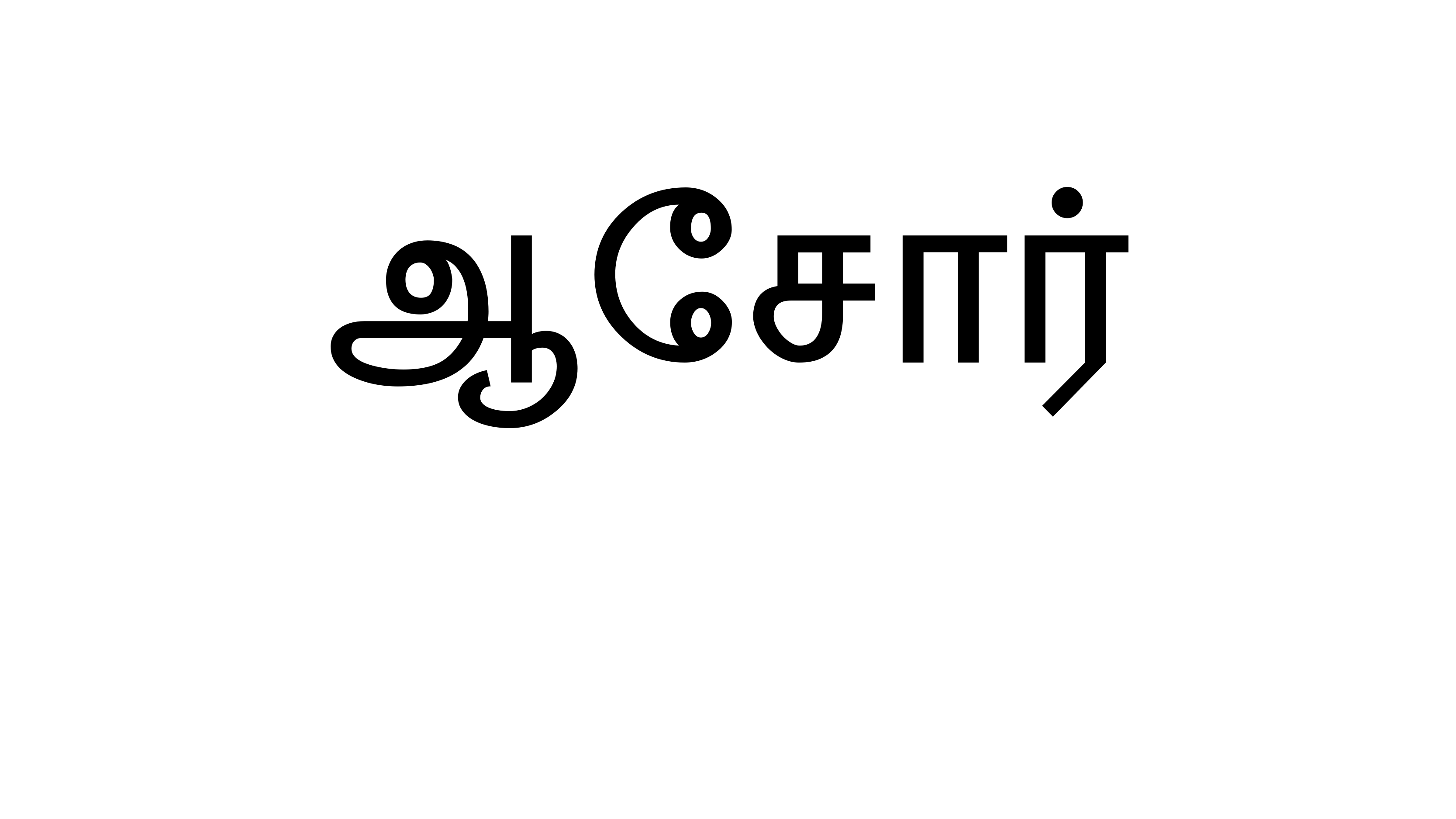}    \\

&&&(eliyuutukkut) & (aacoor)

\end{tabular}
\caption{Examples of improvement due to neural
transliteration. \methodb: incorrect
 prediction
 of \methodb. \methodn: correct prediction obtained with
 neural transliteration.
 \label{tab:freq_vs_neural}}
\end{table}

\subsection{Error analysis}
In our manual error analysis, we found two main types of errors. 

(1) The neural model
generally learns well how to  transliterate the beginning of
a word, but error rates are higher word-internally. For example, the NE ``balak'' is wrongly paired to ``pileeyaam'' instead of ``paalaak'' and ``menna'' is paired to ``meleyaa'' instead of ``meyinaan'' in Tamil.
The neural model has to learn two aspects of
transliteration: transliteration proper (i.e., character
correspondences) and alignment. This type of error indicates
that alignment performance should be improved.
In future work, we plan to explore neural architectures that
more explicitly model the problem as alignment.

(2) For some low-resource languages, the output of \methodb has a high level of noise, so the neural model fails to learn some character
correspondences. In some cases, the output of the neural model
is unrelated to the input. This type of error indicates that the \methodb method should be improved further. 
As shown in Tables \ref{tab:examples}
and  \ref{tab:freq_vs_neural}, low-frequency words contain
more errors. In future work, we plan to adopt an iterative
strategy
that considers gradually more and more named entities, starting with
the most confident ones.

\section{Use cases}\label{sec:applications}
\subsection{Transliteration}
A straightforward application of our named entity resource, as described by \citelanguageresource{wu2018creating}, is 
to create transliteration models. They showed that a 
character-based Moses SMT system  trained over a dataset of named entities extracted
from the Bible (whose performance is lower than our
method's, based on \tabref{mainresults}) performs
better than a Unicode baseline.  
We  now present two additional applications of our named entity resource: 
extending existing multilingual dictionaries and cross-lingual mapping of word embeddings.

\subsection{Extending existing multilingual resources}
BabelNet\footnote{\url{https://babelnet.org/}}
\citelanguageresource{navigli2012babelnet}
 is a multilingual encyclopedic dictionary. It was created
by integrating more than 35 WordNets, covering 500
languages,
and has about
20 million entries. 

We want to show that one can use our resource to enrich BabelNet further.
Since \methodn covers many more languages than BabelNet, we can simply extend
BabelNet by adding more languages like Burarra, North Junín Quechua, and Mian
to it. 
Regarding the languages that BabelNet already supports, we check whether we can add more 
entries exploiting our resource. To this end, for each word pair (English:target-language)
in \methodn, we check whether a translation of the English word exists in BabelNet in the target language. 
Results are depicted in Table \ref{tab:babel}. On average, $27\%$ (i.e., 206 words) of the English words have no correspondence in the target language. 
These are mostly rare words that are difficult to translate without accessing a resource as rich as PBC. 
From a manual investigation, we find that our resource could also help to improve the quality of BabelNet; some translations of the latter are completely incorrect or wrongly written with Latin characters. Examples for Greek are hamor/\begin{otherlanguage*}{greek}εμμώρ\end{otherlanguage*} (emmor), which BabelNet  translates as  \begin{otherlanguage*}{greek}Δείνα\end{otherlanguage*} (Deina), and 
ethan/\begin{otherlanguage*}{greek}εθάν\end{otherlanguage*}, incorrectly transliterated with Latin characters.

\begin{table}[!t]
	\scriptsize
	\centering
	\setlength{\tabcolsep}{3.8pt}
	\resizebox{.45\textwidth}{!}{%
	\begin{tabular}{lrrrr} \hline

		Lang. & CLC-BN & Babel & New NEs & New NEs \% \\
	\hline
		Arb  & 977   & 683   & 294       &  30.1 \\
		Fin  & 979   & 647   & 332       &  33.9 \\
		Ell  & 979   & 658   & 321       &  32.8 \\
		Rus  & 485   & 449   & 36        &  7.4 \\
		Spa  & 979   & 784   & 195       &  19.9 \\
		Swe  & 979   & 684   & 295       &  30.1 \\
		Zul  & 979   & 471   & 508       &  51.9 \\ \hline
		Heb  & 467   & 413   & 54        &  11.6 \\
		Hin  & 467   & 334   & 133       &  28.5 \\
		Kan  & 467   & 299   & 168       &  36.0 \\
		Kor  & 467   & 386   & 81        &  17.3 \\
		Kat  & 368   & 271   & 97        &  26.4 \\
		Tam  & 433   & 318   & 115       &  26.6 \\
		
		\hline
		Jpn    & 979   & 715   & 264       &  27.0 \\
		Zho    & 979   & 698   & 281       &  28.7 \\
		Tha    & 467   & 337   & 130       &  27.8 \\

	\hline
		AVG.  & 715	  & 509   &	206       &	27.2  \\
	\hline
	\end{tabular}}

	\caption{Extension of BabelNet with named entities
        based on our resource. Example (first line, ``Arb''). CLC-BN returns 977
        English-Arabic NE pairs. BabelNet contains Arabic
        translations for 683 of these English NEs, but 294 (30.1\%)
        lack an Arabic translation. Thus we add 294
        English-Arabic NE pairs
that were not covered by BabelNet. \label{tab:babel}}
	
\end{table} 

\subsection{Cross-lingual mapping of word embeddings}

An effective method for creating bilingual word embeddings is to train word embeddings for each 
language independently using monolingual resources and then aligning them 
using a linear transformation \cite{artetxe-etal-2018-robust}. Approaches 
for word embedding alignment
can be grouped into three categories: supervised
\cite{mikolov2013exploiting,lazaridou2015hubness}, semisupervised 
\cite{artetxe2017learning}
and unsupervised  \cite{artetxe-etal-2018-robust,alvarez2018gromov}. 
Supervised approaches require a bilingual dictionary with a few thousand entries to learn the mapping. Semisupervised procedures need a small seed dictionary.
Unsupervised approaches can align word embeddings without 
any bilingual data but, as shown by \newcite{vulic-etal-2019-really},
they are only effective when the two languages 
are similar enough, restricting their applicability.

In this use case, we use our resource as the initial seed
dictionary for semisupervised alignment of word embeddings
for language pairs where unsupervised methods fail. We select
three such language pairs -- English/Japanese,
English/Chinese and English/Tamil -- and show that
VecMap,\footnote{\url{https://github.com/artetxem/vecmap}}
a semisupervised method, can successfully employ
our NE resource
to align these languages.  VecMap implements
the method proposed by \newcite{artetxe-etal-2018-robust},
which is a state-of-the-art method for unsupervised
cross-lingual word embedding mapping.  It creates an initial
set of word pairings based on the distribution of words in
their similarity matrix. Then it employs a self-learning
method to improve the mapping iteratively.

We evaluate the embeddings on the 
Bilingual Lexicon Induction (BLI) task and the gold dataset provided by MUSE \citelanguageresource{Conneau2018WordTW}.
We use Wikipedia fastText embeddings \citelanguageresource{bojanowski-etal-2017-enriching} 
as monolingual input vectors and report precision at one (P@1) for the unsupervised and semisupervised approaches in Table \ref{tab:bli}. While the fully unsupervised method
 fails to align these languages, the semisupervised
 approach based on our resource has much better results confirming that our NE resource can be effectively used as seed data.

\begin{table}[!t]
	\centering
	\small
	\begin{tabular}{lrrr} \hline
		 	 		   & Eng-Jpn   & Eng-Tam & Eng-Zho   \\ \hline
		Unsupervised   & 0.0     & 0.0   & 0.0     \\
		Semisupervised & 30.43   & 14.4  & 30.1

	\end{tabular}
\caption{P@1 BLI results with unsupervised VecMap compared
to semisupervised VecMap, which uses
our NE resource for initialization
\label{tab:bli}}
\end{table}

\section{Resource}
We release a resource of named entities for 1340 languages, 1134 of which are lowest-resource.\footnote{Our NEs resource is freely available at \url{http://cistern.cis.lmu.de/ne_bible/}}
The resource mainly contains people and location NEs.
The total number of NEs is 674,493, so there are 503 NEs per language on average with at least 300 NEs in 95\% of the languages.
The three best represented language families \citelanguageresource{wals} are Austronesian, Niger-Congo and Indo-European. 
However, our coverage broadly includes all major areas of linguistic diversity,
including Amazonian (e.g., Kaingang), African (e.g., Sango) and Papua New Guinea (e.g., Saniyo-Hiyewe).

\section{Conclusion}
We presented \methodn, a new method that identifies named
entity correspondences and trains a neural transliteration
model on them. \methodn does not need any other bilingual
resources beyond the parallel corpus nor a word aligner or
seed data.  We showed that it outperforms prior work on
silver data and human-annotated gold data.  We created a new
NE resource for 1340 languages by applying \methodn to the Parallel Bible
Corpus
and
illustrated its utility by demonstrating good performance on
two downstream
tasks: knowledge graph augmentation and bilingual lexicon
induction.

\smallskip

\textbf{Acknowledgments.}
This work was funded by the European Research Council
(grant \#740516) and
the German Federal Ministry of Education and
Research (BMBF, grant \#01IS18036A).

\section{Bibliographical References}\label{reference}

\bibliographystyle{lrec2022-bib}
\bibliography{lrec2022-example,anthology,custom}

\section{Language Resource References}
\label{lr:ref}
\bibliographystylelanguageresource{lrec2022-bib}
\bibliographylanguageresource{languageresource}

\appendix

\section{Reproducibility Information} \label{sec:repr}
We run our method on up to 48 cores of Intel(R) Xeon(R) CPU E7-8857 v2 with 1TB memory and a single GeForce GTX 1080 GPU with 8GB memory. 
\methodn  is implemented in Python and takes approximately 2
minutes to run for one language. 
The neural model is implemented in PyTorch and has one encoder and one decoder layer (batch size 16, hidden layer size 32, learning rate 0.01, dropout 0.4, 24K parameters).
We use \newcite{luong2015effective}'s attention. Each training of the neural transliteration model requires at most 10 minutes.
SimAlign \cite{sabet2020simalign} alignments are obtained using multilingual BERT \cite{devlin-etal-2019-bert}. We use subword alignments and the forward alignment to ensure that all English NEs are aligned. 
Eflomal \cite{ostling2016efficient} alignments are obtained with default parameters and the forward alignment.
The Jaro distance is calculated using the Python library textdistance.\footnote{\url{https://pypi.org/project/textdistance/}}

For the cross-lingual word alignment experiment we used the 
latest VecMap code available in its git repository\footnote{commit ID:\\
	\texttt{b82246f6c249633039f67fa6156e51d852bd73a3}} (no snapshot is available).
We ran it using the $<--unsupervised>$ and $<--semi\_supervised>$ switches. 
All other parameters are left as their default value.
The monolingual word alignments are downloaded from fastText's
official website.\footnote{\url{https://fasttext.cc/docs/en/pretrained-vectors.html}}

\end{document}